%% file: main.tex
\documentclass[10pt,twocolumn,letterpaper]{article}

\usepackage{cvpr}
\usepackage{times}
\usepackage{epsfig}
\usepackage{graphicx}
\usepackage{amsmath}
\usepackage{amssymb}
\usepackage{bbm}
\usepackage{subcaption}
\usepackage{multirow}
\usepackage{adjustbox}
\usepackage[dvipsnames]{xcolor}
\usepackage{tabu}
\usepackage{floatrow}
\usepackage{booktabs}
\usepackage{makecell}
\usepackage{epigraph}

\usepackage{algorithm}
\usepackage{algorithmic}

\usepackage[pagebackref=true,breaklinks=true,letterpaper=true,colorlinks,bookmarks=false]{hyperref}

\newcommand{\MYMODULE}{feedback module }
\newcommand{\E}{\mathcal{E}}
\newcommand*{\ShowNotes}{}

\definecolor{darkred}{rgb}{0.7,0.1,0.1}
\definecolor{darkgreen}{rgb}{0.1,0.7,0.1}
\definecolor{cyan}{rgb}{0.7,0.0,0.7}
\definecolor{dblue}{rgb}{0.2,0.2,0.8}
\definecolor{maroon}{rgb}{0.76,.13,.28}
\definecolor{burntorange}{rgb}{0.81,.33,0}

\ifdefined\ShowNotes
  \newcommand{\colornote}[3]{{\color{#1}\bf{#2: #3}\normalfont}}
\else
  \newcommand{\colornote}[3]{}
\fi

\cvprfinalcopy 

\def\cvprPaperID{2388} 

\ifcvprfinal\fi
\begin{document}

\title{Adversarial Feedback Loop}
\author{Firas Shama\qquad Roey Mechrez \qquad Alon Shoshan \qquad Lihi Zelnik-Manor\\
Technion - Israel Institute of Technology\\
{\tt\small \{shfiras@campus,roey@campus,shoshan@campus,lihi@ee\}.technion.ac.il}}
\maketitle

\begin{abstract}
Thanks to their remarkable generative capabilities, GANs have gained great popularity, and are used abundantly in state-of-the-art methods and applications. In a GAN based model, a discriminator is trained to learn the real data distribution. To date, it has been used only for training purposes, where it's utilized to train the generator to provide real-looking outputs. In this paper we propose a novel method that makes an explicit use of the discriminator in test-time, in a feedback manner in order to improve the generator results. To the best of our knowledge it is the first time a discriminator is involved in test-time. We claim that the discriminator holds significant information on the real data distribution, that could be useful for test-time as well, a potential that has not been explored before. 

The approach we propose does not alter the conventional training stage. At test-time, however, it transfers the output from the generator into the discriminator, and uses feedback modules (convolutional blocks) to translate the features of the discriminator layers into corrections to the features of the generator layers, which are used eventually to get a better generator result. Our method can contribute to both conditional and unconditional GANs. As demonstrated by our experiments, it can improve the results of state-of-the-art networks for super-resolution, and image generation.
\end{abstract}

\input{introduction.tex}

\input{RelatedWork.tex}

\input{method.tex}

\input{experiments.tex}

\input{conclusion.tex}

\paragraph{Acknowledgements} This research was supported by the Israel Science Foundation
under Grant 1089/16 and by the Ollendorf foundation.

{\small
\bibliographystyle{ieee}
\bibliography{egbib}
}

\end{document}

%% file: introduction.tex
\begin{figure}[t]
    \centering
  \includegraphics[width=0.98\linewidth ]{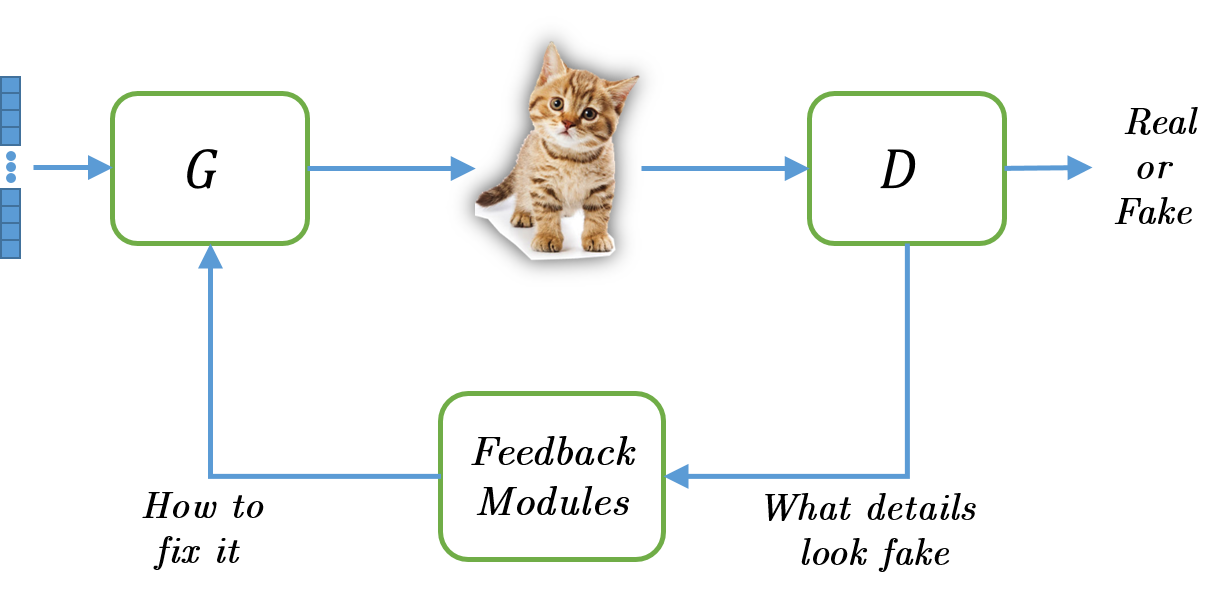}
  \caption{\textbf{The adversarial feedback loop}: Classic GAN is composed of two components: the generator ($G$) and the discriminator ($D$). In this setting, the information flow is done purely by back-propagation, during training. We propose adding a third component -- the \textit{feedback module} that transmits the discriminatory spatial information to the generator in a feedback manner at inference time.}
  \label{fig:first}
\end{figure}%

\section{Introduction}
\label{sec:intro}
Adversarial training \cite{goodfellow2014generative} has become one of the most popular tools for solving generation and manipulation problems. For example in image generation~\cite{goodfellow2014generative, radford2015unsupervised}, super-resolution~\cite{ledig2017photo}, image-to-image transformation~\cite{isola2017image,zhu2017unpaired}, text-to-image~\cite{reed2016generative} and others. Common to all these works is the discriminator--generator information flow -- via a loss function. That is, the generator output images are fed into the discriminator which produces a `real-fake' score for each image in terms of a pre-defined loss function.
This score is back-propagated to the generator through gradients. 

Recent research in the GAN field discusses the design of the loss function and regularization terms. For example, the basic cross-entropy loss~\cite{goodfellow2014generative}, Wasserstein distance~\cite{arjovsky2017wasserstein}, spectral normalization~\cite{miyato2018spectral} or relativistic discriminator~\cite{jolicoeur2018relativistic}. 
This discussion has contributed significantly to the advancement of GANs and using a discriminator has become highly effective. 
To date, after training the discriminator is forsaken and the deep understanding of the data distribution is lost.  
This seems wasteful to us, hence, we seek a way to enjoy the discriminator also during test-time.
In addition, encapsulating the discriminator information into a single score looses the spatial understanding of which regions are more `real' and which are considered `fake'. 
In the current scheme only limited spatial information flows with the back-propagation because the gradients are averaged over each batch.


In this paper we propose a different approach that explicitly exploits the discriminator's activations, in test-time, in order to improve the generator output. 
We propagate the discriminator information back to the generator utilizing an iterative feedback loop, as illustrated in Figure~\ref{fig:first}. 
The overall framework is as follows: 
We start with classic training of the generator and discriminator.
Then, at test-time, the generator produces an output image which fed into the discriminator in order to compute its feedback.
The discriminator activations are fed into a third module which we name --  \emph{feedback module}. 
The goal of this module is to convert the discriminator activations into `corrections' which can then be added to the original generator activations. 
We repeat this process iteratively until convergences ($1$-$3$ iterations). 

\begin{figure}
\centering
\begin{tabular}{ccc}
   \rotatebox[origin=l]{90}{\centering \tiny{DCGAN}}  &      \rotatebox[origin=l]{90}{\centering \cite{radford2015unsupervised}}  &   \includegraphics[width=0.9\linewidth]{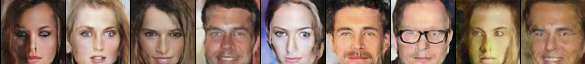}\\
   \rotatebox[origin=l]{90}{\centering \tiny{DCGAN}}  &       \rotatebox[origin=l]{90}{\centering \footnotesize{+AFL}}  &    \includegraphics[width=0.9\linewidth]{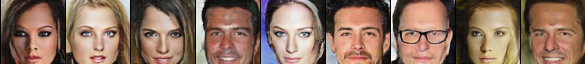} \\
  &
   \rotatebox[origin=l]{90}{\centering \footnotesize{$\sqrt{\mbox{diff}}$}}  &    \includegraphics[width=0.9\linewidth]{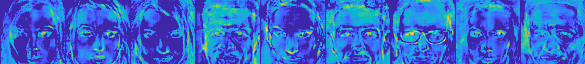}
\end{tabular}
\vspace{-0.2cm}
  \caption{\textbf{Contribution of AFL:} Face generation results of (top) DCGAN~\cite{radford2015unsupervised} and (middle) integrating DCGAN within the proposed AFL framework. Faces generated by DCGAN+AFL are sharper and show fewer artifacts. (bottom) The differences between the images generated without and with AFL highlight that AFL corrects the spatial regions that suffer from artifacts, for example, the cheek of the left-most face, and the eyes of the second from left.
  }
  \label{fig:CP_face}
\end{figure}

The main contributions of the Adversarial Feedback Loop (AFL) are two-fold.
First, to the best of our knowledge, our novel adversarial feedback loop is the first use of the discriminator at test-time. 
Second, our scheme makes the spatial discriminative information accessible to the generator, allowing it to `correct' artifacts and distortions thus producing higher quality images.
A few motivational examples are presented in Figure~\ref{fig:CP_face}, where it can be seen that our pipeline takes images produced by the generator and corrects those regions that suffered from artifacts.

Experiments on the CIFAR-10~\cite{krizhevsky2009learning} and CelebA~\cite{liu2015faceattributes} data-sets for the task of unsupervised image generation show that combining AFL with state-of-the-art GAN methods improves the Inception Score (IS)~\cite{salimans2016improved} which is evident also qualitatively as better visual quality.
In addition, when integrated with ESRGAN~\cite{wang2018esrgan}, a state-of-the-art method for super-resolution, AFL can further improve the results; it achieves higher Perceptual Index ~\cite{blau20182018} and lower RMSE, making the results more visually appealing and more trustworthy to the ground truth.


%% file: RelatedWork.tex
\section{Related Work}
\label{sec:relatedWork}
The idea of exploiting the discriminator features served as motivation to some previous works. 
The GAN-based methods~\cite{wang2017high,larsen2015autoencoding} proposed to use a loss based on features extracted from the discriminator layers.
They compared the discriminator features of fake image to the discriminator features of real images, in a similar manner to the renowned perceptual loss~\cite{dosovitskiy2016generating}.
In all those methods, the utility of the discriminator layers was limited to training-time, which is different from our suggestion to enjoy its benefits also in test-time.

The concept of feedback has already made its way into the training framework of several previous works that exploit the iterative estimation of the output aiming at better final results. 
In~\cite{oberweger2015training} a feedback loop was trained for hand pose estimation, in~\cite{Carreira_2016_CVPR} feedback was used for human pose estimation, while~\cite{li2016iterative} proposed to use the feedback for the problem of instance segmentation. \cite{zamir2017feedback} suggest a general feedback learning architecture based on recurrent networks, that can benefit from early quick predictions and from a hierarchical structure of the output in label space. 
An interesting solution for the task of video frames prediction was presented in~\cite{lotter2016deep}, that introduced an unsupervised recurrent network that feeds the predictions back to the model.
Feedback was also used for super-resolution by ~\cite{haris2018deep} that suggest a network that uses the error feedback from multiple up and down-scaling stages. 

To the best of our knowledge, none of these previous method has proposed applying the concept of feedback in the framework of GAN.
To place our proposed framework in context with the terminology common in works discussing feedback paradigms, one can think of the generator as a `predictor', the discriminator as an `error estimator' and feedback modules close the loop and convert the errors from the discriminator feature space to the generator feature space.

%% file: method.tex
\input{fig_training_arch.tex}

\section{Method}
\label{sec:method}

In this section we present our AFL framework.
As discussed in the introduction, all current methods use the discriminator for adversarial training only.
We wish to change that and explicitly use the knowledge that the discriminator has gathered also during test-time.
This way the discriminator can "leak" information to the generator by providing feedback on the generator's failures and thus assisting the generator in fixing them. 
We designed a solution that is generic so that it can be integrated with any GAN based network.


\subsection{Framework Overview}
Given a generator-discriminator architecture, $G$ and $D$ respectively, we denote their layers by $\{g^l\}_{l=1}^n$ and $\{d^l\}_{l=1}^n$, where $l$ is the layer index.
These layers (or some) are connected via feedback modules $\{f^l\}_{l=1}^n$\footnote{Each of which consists of two convolutional layers: \\({$CONV\!-\!BN\!-\!RELU\!-\!CONV\!-\!BN$})}.
The input to each feedback module is the activation map of the corresponding layer of the discriminator $\theta^l=d^l(\theta^{l-1})$.
The output of each feedback module is added to the corresponding activation map of the generator $\phi^l=g^l(\phi^{l-1})$, thus forming a \emph{skip-connection}, such that the generator activation maps change to:
\begin{equation}
    \phi^{l+1}=g^{l+1}\left(\phi^l+f^l(\theta^l)\right).
\end{equation}
See Figure \ref{fig:framework}{\color{red}(a)} for illustration. 
Each feedback module is further associated with a scalar parameter $\alpha^l$ that multiplies its output.
Setting $\alpha^l=0$ deactivates the $l$'th module altogether, while $\alpha^l\neq0$ tunes the contribution of the feedback.

The basic \MYMODULE connects between equally-sized activation layers of the discriminator and the generator. 
We also suggest a slightly more complex form, where the \MYMODULE\!s are given as input the activation maps of both the discriminator and the generator (concatenated, noted as $[\cdot,\cdot]$), as illustrated in Figure~\ref{fig:framework}{\color{red}(b)}, such that the generator activation maps change to:
\begin{equation}
    \phi^{l+1}=g^{l+1}\left(\phi^l+f^l([\theta^l,\phi^l])\right).
\end{equation}

\subsection{Training} 
The training scheme consists of two phases.
The first phase is identical to the common practice in training GANs. 
The feedback modules are inactive and we apply standard adversarial network training, in which the generator, $G$, and the discriminator, $D$, are trained according to the selected base method.
The outcome is a trained generator and a trained discriminator that can differentiate between real images and fake images produced by the generator.

The second training phase is where we activate the feedback modules and train them. 
This is done while freezing the generator $G$, but allowing the discriminator $D$ to keep updating.
This way the feedback modules learn to correct the generator results in order to improve them based on the feedback given from the discriminator. 
Since the output from the generator improves, we must allow the discriminator to continue and refine its weights.

We next write in further detail the steps of the second phase of the training:

\paragraph{First iteration $t=0$}
Given input $x$ (e.g., a random vector), the generator produces an initial output image $y_0=G(x)$ that is fed into the discriminator. 

\paragraph{The $t>0$ iteration}
We set $\alpha^l=1$ and use the following update equation:
\begin{equation}
    y_{t+1} = G\left(x,\E(y_t)\right)
\end{equation}
where $\E$ aggregates the output of all the feedback modules:
\begin{equation}
    \mathcal{E}(y_t) = \{f^l\left(\theta^l(y_t)\right)\}_{l=1}^{n}
\end{equation}
or
\begin{equation}
    \mathcal{E}(y_t) = \{f^l\left([\theta^l(y_t),\phi^l(y_{t-1})]\right)\}_{l=1}^{n}
\end{equation}
depending on which feedback module type is used.
In practice, in almost all our experiment two iterations sufficed, i.e. until we get $y_1$.



\paragraph{The Objective}
The feedback modules are trained with the same objective as the baseline generator (e.g. cross entropy, Wasserstein distance, etc.), while replacing every instance of the term $G(x)$ with the term $G(x,\E)$.

\subsection{Testing}
At test-time we freeze the entire network including the generator, the discriminator and the feedback modules.
The activation levels of the feedback modules are tuned by setting $\alpha^l=\alpha$, i.e.
\begin{equation}
   y_{t+1} = G\left(x,\alpha\cdot\E(y_t)\right).
\end{equation}
Typically the impact of the corrections from the feedback modules needs to be attenuated and we have found empirically that best results are obtained when $0.1\leq\alpha\leq0.2$.
This way only the stronger corrections really contribute.

Note, that because of the batch-norm layer in each feedback module, its output signal is forced to be with the same strength (variance) as the generator features, such that multiplying the output by a small $\alpha$ is sufficient in order to preserve the original features, and marginally correct them.
Unless otherwise specified, in all our experiments we used the value of $\alpha=0.2$.

In principle the test-time process could also be repeated iteratively, however, we have found that it suffices to run a single iteration to achieve satisfactory results.

%% file: fig_training_arch.tex
\begin{figure*}
\setlength{\tabcolsep}{20pt}
\begin{tabular}{cc}
    \includegraphics[height=5cm]{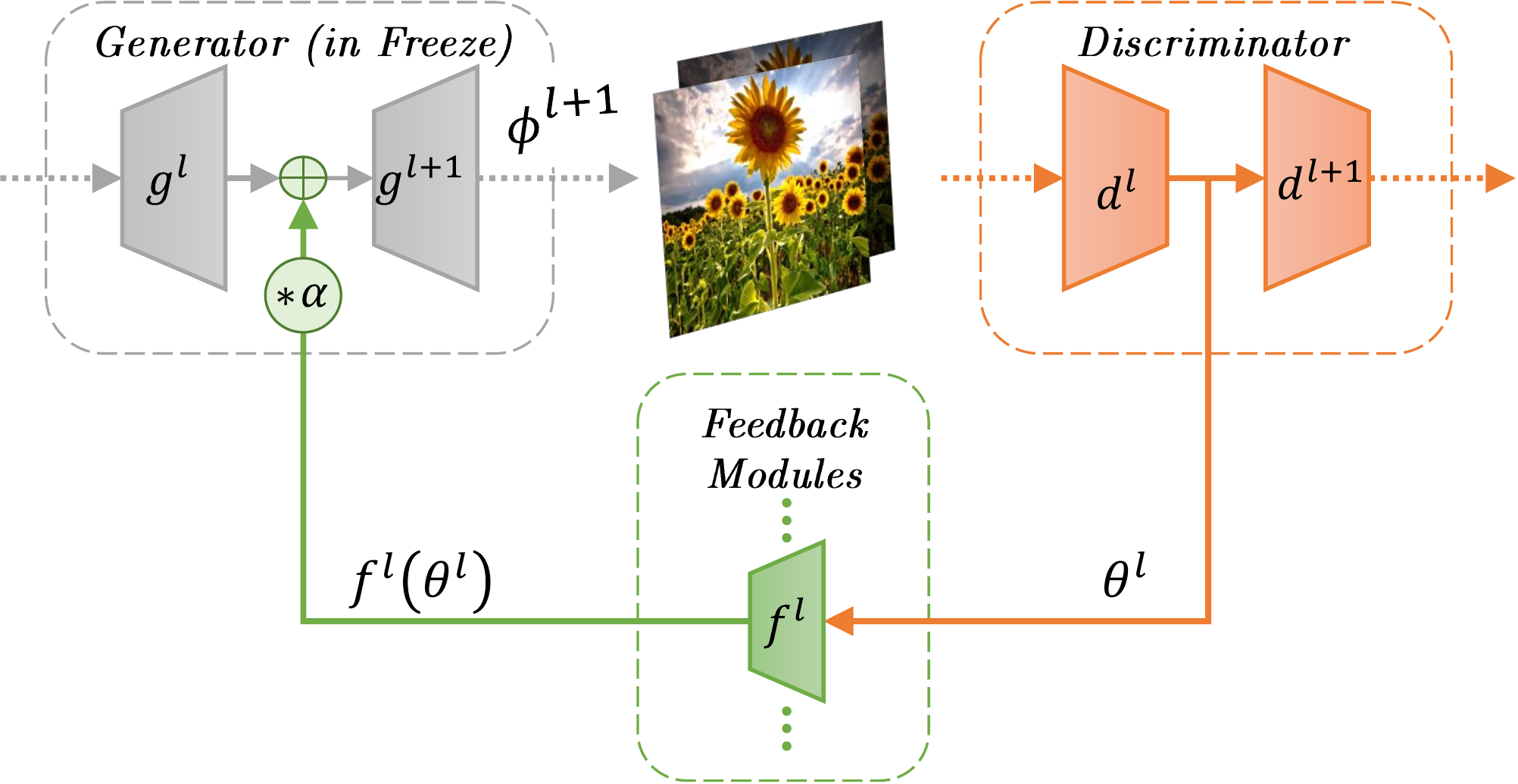} 
     & \includegraphics[height=5cm]{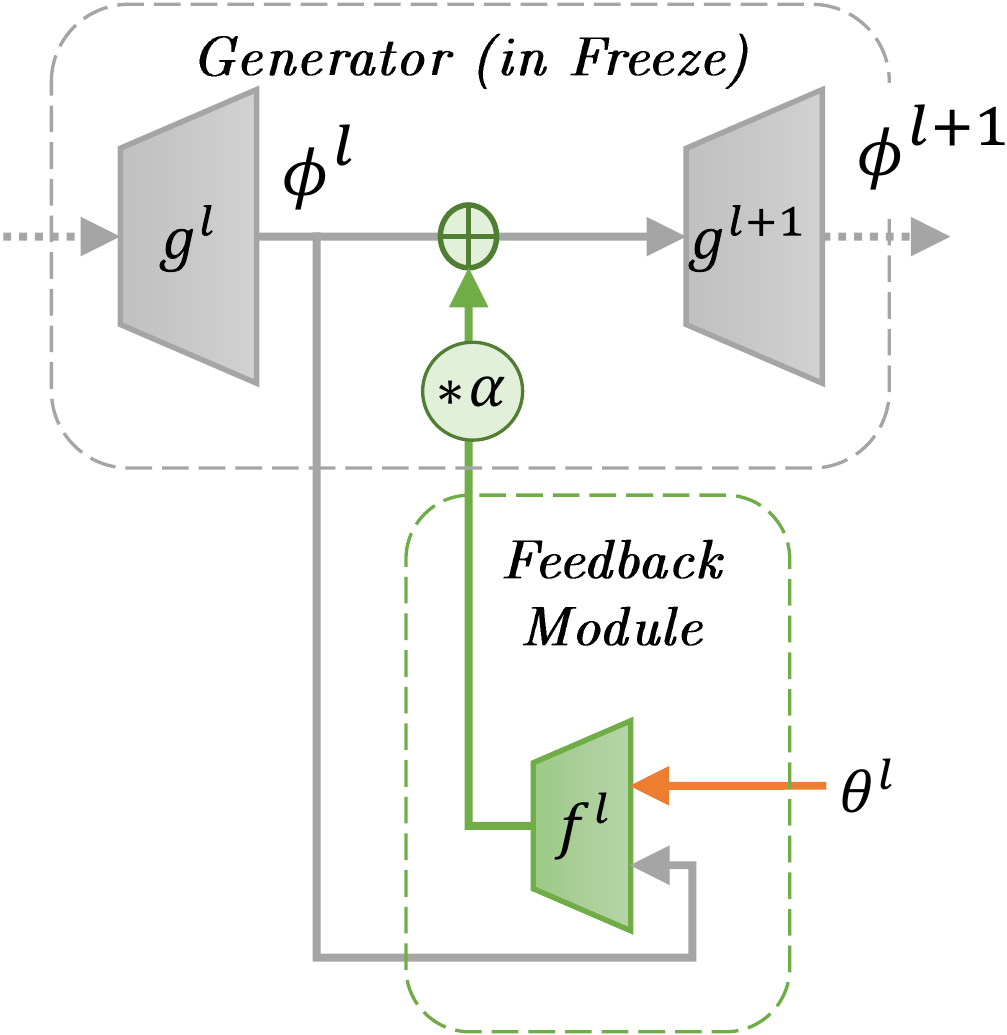} \\
    (a) Generator--Discriminator--Feedback module pipeline   & (b) Dual-input feedback module
\end{tabular}
\caption{\textbf{The feedback framework:} (a) The proposed feedback module passes information from the discriminator to the generator thus ``learning'' how to correct the generated image in order to make it more real in terms of the discriminator score. 
(b) It is also possible to let the feedback module consider both the features of the discriminator and the features of the generator.}
\label{fig:framework}
\end{figure*}

%% file: experiments.tex
\section{Experiments}
In this section we present experiments conducted on several data-sets, tasks and networks that demonstrate the contributions of our method. 
In all cases we add our AFL to existing methods while adopting their training details: the architecture, the loss function, the normalization term and the hyper-parameters. 
Our modifications are reserved to the second training phase, in which we train the \MYMODULE and to the testing phase, where we use the \MYMODULE during generation. 


\input{toy.tex}

\input{cifar10.tex}

\input{celebA.tex}
\input{SR.tex}

%% file: toy.tex
\subsection{Empirical analysis on a simple 2D case}
\label{sec:toy_prob}
Before diving into applications, we first perform
empirical evaluation of AFL in a simple, yet illustrative, scenario. The goal here is to show that AFL is effectively able to utilize the discriminator information in order to improve the generated results. 

The scenario we chose is generation of 2D coordinates that lie on a `Swiss roll'.
The generator gets a random input points $x\!\in\!{\rm I\!R}^2$ and is trained to generate points $y\!\in\!{\rm I\!R}^2$ that fit into the `real' data distribution, represented by the discriminator.
The discriminator is trained to classify each sample as `real' or `fake' according to a given ground-truth swiss-roll data distribution.

As architecture, we chose both the generator and the discriminator to consist of a sequence of four fully-connected layers.
For the feedback we used a single module, that corrects the input of the last layer of the generator. 
The objective we used was the WGAN-GP~\cite{gulrajani2017improved} adversarial loss. Please refer to the supplementary for implementation details.
As baseline model the generator and the discriminator were trained for $8K$ iterations. Then we froze the generator, added the feedback module and trained it with the same discriminator for another $8K$ iterations.

\begin{figure}
\centering
\begin{tabular}{c}
\includegraphics[width=.99\linewidth]{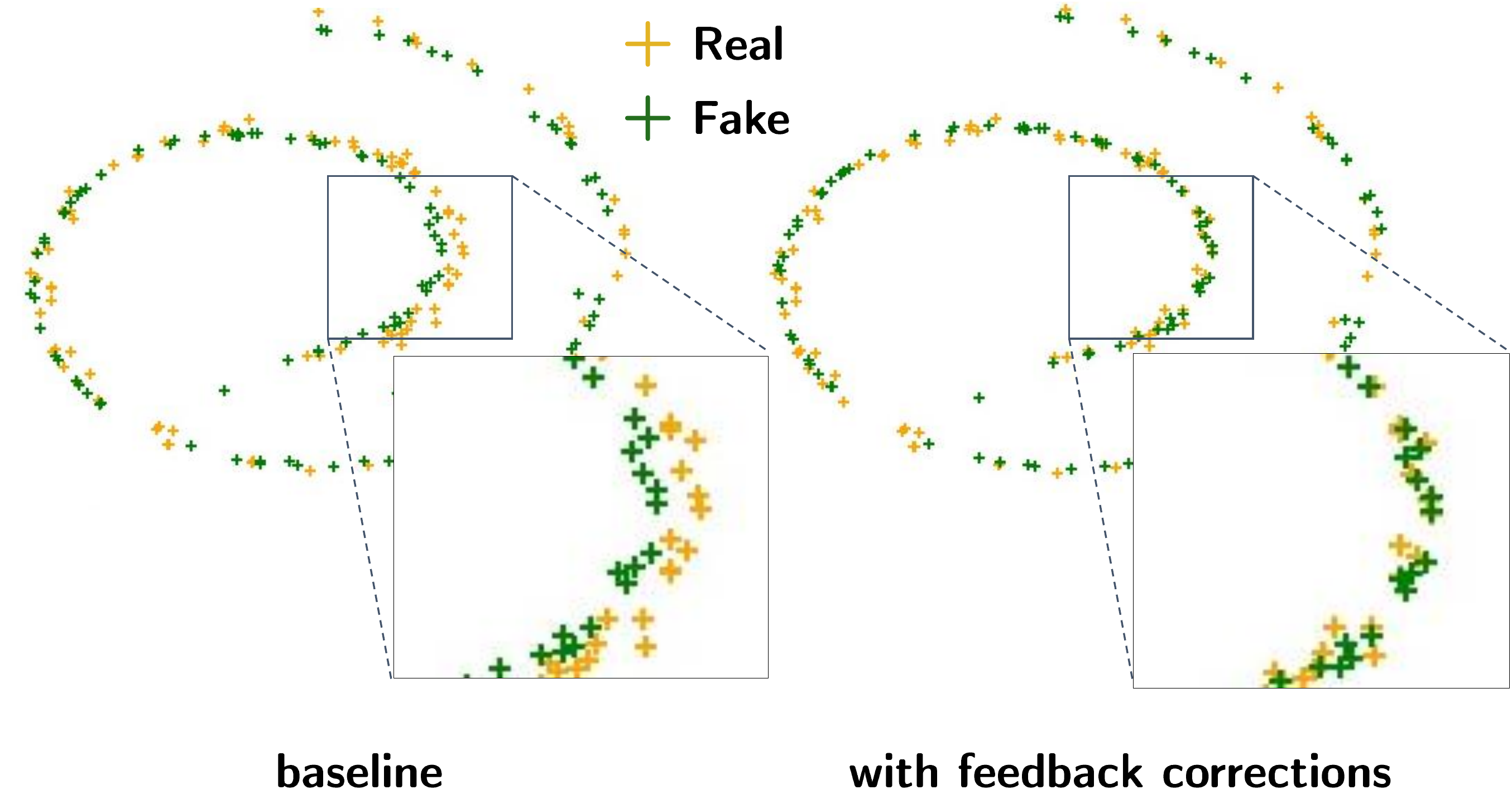}\\
(a) Test input variance = training input variance \\
\\
\includegraphics[width=.99\linewidth]{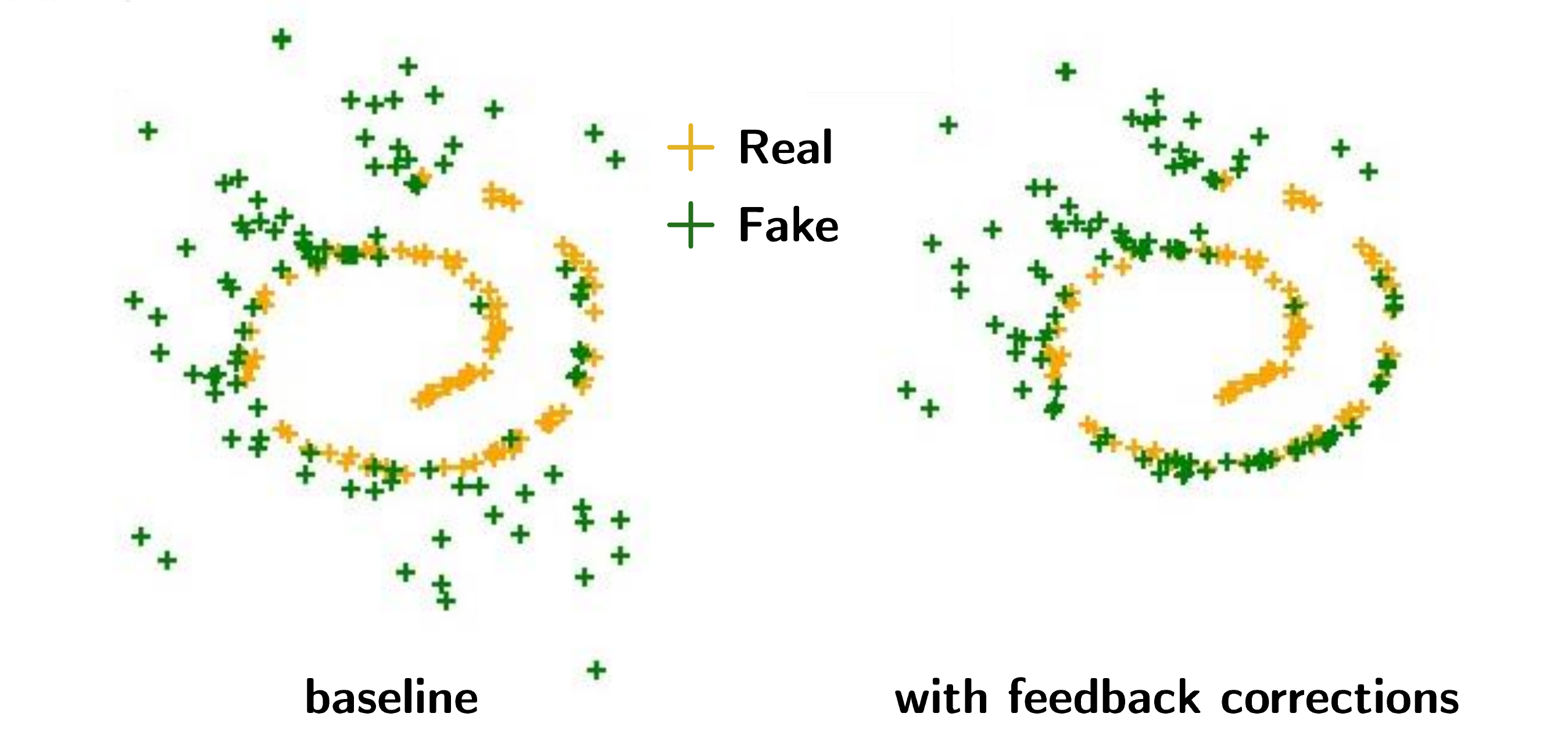}\\
(b) Test input variance = \textbf{$5\times$} training input variance \\
\end{tabular}
\caption{\textbf{Swiss-roll generation:} 
Results of generating points that lie on a swiss-roll.
(a) When the variance of the random input during test-time matches the variance of the inputs used for training the baseline generator does a decent job. Adding AFL corrects the small inaccuracies and yields a distribution almost identical to the real one.
(b) When the variance of the random input is increased, the baseline generator fails. Conversely, using AFL has still succeeds to some extent to reproduce the swiss-roll distribution.
}
\label{fig:toy_iters}
\end{figure}

Our results are presented in Figure~\ref{fig:toy_iters}.
It can be observed that the baseline generator succeeds to generate samples close to the real data distribution, however, using the proposed AFL pipeline improves the generation accuracy and results in a distribution that is much closer to the real one.
The AFL module identifies that inaccuracies in the generated points, corrects them, and leads to better results.

%% file: cifar10.tex
\input{fig_cifar10_3methods.tex}

\subsection{Image generation on CIFAR-10} 
\label{sec:cifar}
As a second test-case we chose the task of unsupervised image generation on CIFAR-10 dataset.
The goal is to generate $32\times32$ images given as input vectors with normally distributed elements, $x\in \mathbb{R}^{128}$.
The generated images should match the diversity and quality of the training dataset used by the discriminator. 
We choose the commonly used Inception Score (IS)~\cite{salimans2016improved} as a quality measure (higher is better) of the generated images, as it has been shown to be strongly correlated~\cite{huang2018empirical,borji2018pros} with generated data quality.
As accepted, we generate $50K$ samples and compute the mean IS. 

Many methods have been proposed before for this task.
In order to demonstrate the generality of our approach to the choice of architecture and loss, we performed experiments with three different baseline methods.
The first is WGAN-GP~\cite{gulrajani2017improved}\footnote{Used pyTorch implementation \href{https://github.com/caogang/wgan-gp}{github/caogang/wgan-gp}} with a reported IS of $6.68$.
The second is SN-GAN~\cite{miyato2018spectral}\footnote{Used pyTorch implementation \href{https://github.com/christiancosgrove/pytorch-spectral-normalization-gan}{github/christiancosgrove/sn-gan}} with a reported IS of $7.42$. 
Both of these methods are GAN based, built on standard CNN architecture, and try to comply to Lipschitz~\cite{arjovsky2017wasserstein} constraints by limiting the weights/gradients of the discriminator.
In selecting a third baseline we wanted to further show that AFL is effective even when no smart or complex normalization/constraints are applied.
Therefore, we choose a standard DCGAN \cite{radford2015unsupervised} network. 

The AFL included a single feedback module that connected between the intermediate layer features with spatial size of $8\times8$. 
For the SN-GAN network, we used the dual-input type of feedback module as it gave better results, while for the other two networks we used the basic type of single input.

Each baseline model was trained according to its recommended scheme. Once the baseline network is trained and perform close to the reported IS, we freeze the generator and train the \MYMODULE and the discriminator.

Figure~\ref{fig:Inception} compares the performance of the three baseline networks, to those with AFL.
It is evident that in all three cases AFL yields a significant improvement in the mean score.
We further present the performance of several other recent methods, just to give a broader perspective on the performance contribution by AFL.

\paragraph{Sanity Check}
To validate that the \MYMODULE indeed learns to harness the feedback from the discriminator we performed two sanity-check experiments.

In the first one we used the AFL pipeline, with WGAN-GP as baseline, but rather than using the feedback from the discriminator, we fed the AFL with normally distributed input. 
the results was a very low IS of $3.44$ on CIFAR-10 (compared to $7.07$ in the proposed AFL setup).
This result shows that the feedback module develops a strong dependency on the discriminator feedback provided in test-time.

As a second sanity check we performed a similar experiment, this time using as feedback the activation maps of the discriminator, albeit, computed on a different (wrong) image from the batch.
Interestingly, this abuse of the solution results in IS of $5.76$, which is better than the score of the first sanity check, but still far behind the score when using the correct feedback.
This stands to show that the \MYMODULE learns to create corrections based on the discriminator feedback on the specific image.

%% file: fig_cifar10_3methods.tex
\begin{figure}
\centering
\setlength{\tabcolsep}{.3em}
\begin{tabular}{c}
 \includegraphics[width=0.9\linewidth]{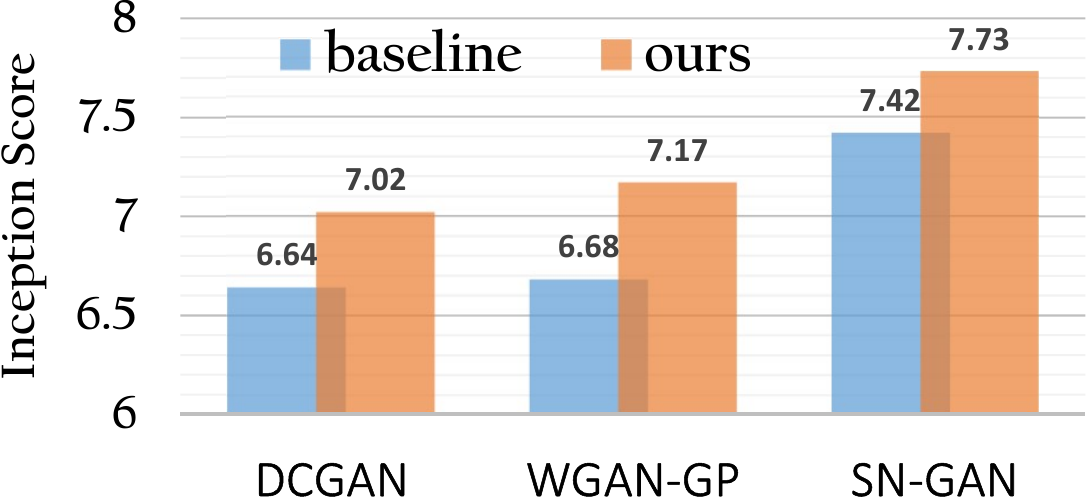} \\
 \\
 \renewcommand{\arraystretch}{1.1}
 \begin{tabu}{@{}lcccc@{}}
      \toprule
	\textbf{Method} &&  \textbf{Baseline} &&  \textbf{Baseline+AFL}\\
    \midrule
        DCGAN~\cite{radford2015unsupervised}* && $6.64\pm.14$ && $\textbf{7.02}\pm.06$ \\
        WGAN-GP\cite{gulrajani2017improved} && $6.68\pm.06$ && {$\textbf{7.17}\pm.05$}\\
        SN-GAN \cite{miyato2018spectral} && $7.42\pm.08$ && {$\textbf{7.73}\pm.05$} \\
        \bottomrule
        *{as reported in \cite{yang2017lr}}
    \end{tabu}
\end{tabular}
\caption{\textbf{Image generation with AFL:} Inception Scores of recent methods in unsupervised image generation on CIFAR-10. Combining our proposed AFL (in orange) with any one of three baselines improves the scores significantly.}
\label{fig:Inception}
\end{figure}

%% file: celebA.tex
\input{fig_FIDvsAlpha.tex}

\subsection{Face generation on CelebA}

\begin{figure*}
  \includegraphics[width=\linewidth]{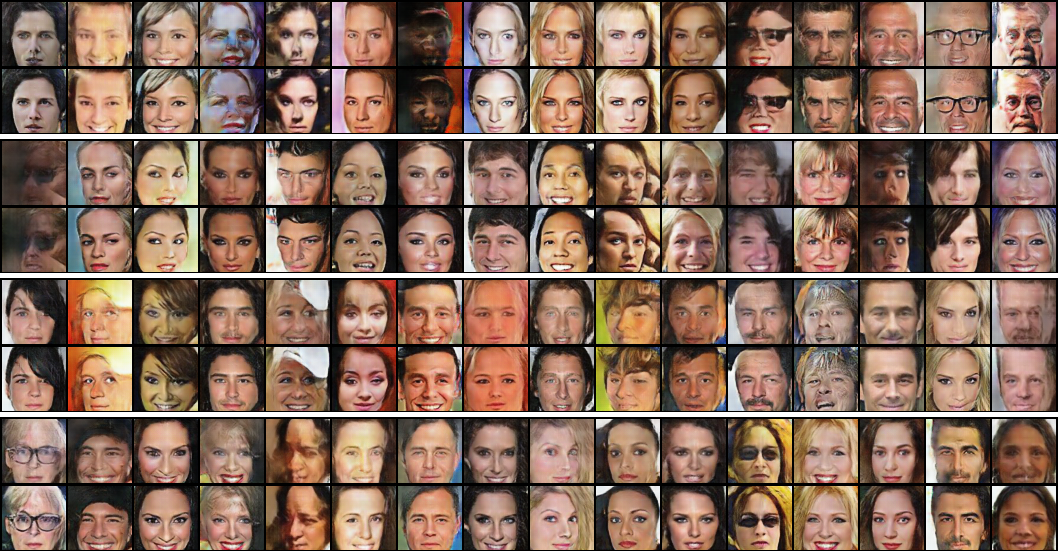}
  \caption{\textbf{Face generation on CelebA:} We compare DCGAN~\cite{radford2015unsupervised} baseline (odd rows) with ours, DCGAN+AFL (even rows). Faces generated with AFL show significantly fewer artifacts, making clear the advantage of using AFL.}
  \label{fig:batch_face}
\end{figure*}

\input{fig_disabling_modules.tex}

As a third task we chose unsupervised face image generation on CelebA dataset~\cite{krizhevsky2009learning}, because it allows both quantitative evaluation as well as qualitative evaluation through visualizing the generated faces.
The output here consists of face image of size $64\times64$, thus a human observer can easily detect distortions and artifacts (unlike CIFAR-10 where visual inspection is not considered effective).

As baseline network we adopted DCGAN~\cite{radford2015unsupervised}.
For feedback we used the dual-input option with four feedback modules, each with two convolutional layers.
First phase training followed the recommended setup until convergence and obtaining results comparable to those originally reported in~\cite{radford2015unsupervised}. 
The second phase trained the feedback modules with the discriminator, for the same number of epochs as in the first phase.

To quantitatively evaluate the contribution of AFL over the baseline DCGAN network, we use the \emph{Fr\'echet inception distance} (FID)~\cite{heusel2017gans} to measure the distance between the distribution of the real data and the generated data.
The real distribution is estimated based on $50$K samples.
We performed an evaluation on several values of $\alpha$, as presented in Figure~\ref{fig:FIDvsAlpha}, all of them improve the baseline result ($\alpha=0$) which has FID of $24.17$. 
The best quantitative result was achieved with $\alpha=0.5$ yielding to FID of $17.32$.
Arguably, in qualitative inspection, often the best looking results were with $\alpha=0.2$, hence, we present in Figure~\ref{fig:batch_face} some of these images.
In almost all samples we can see a clear quality improvement brought in by AFL (more on the supplementary).

To provide further insight on why AFL is useful we performed another experiment, in which we study the contribution of each module of AFL.
Specifically, we disabled ($\alpha^l=0$) all the feedback modules except for the one whose impact we want to study, and repeated that for all the modules.
Results are presented in~\ref{fig:disabling_modules_exp}.
As could be speculated, the shallowest module has the strongest correction impact.
Since it has the biggest ``receptive field'', it learned to correct high- mid- level errors, such as the head shape.
The last module learned to correct low-level features such as colors and skin-tone.
Finally, the intermediate modules learned to correct eye gaze, smile, nose shape etc.

\paragraph{Feedback Switching}

\begin{figure}
  \includegraphics[width=0.99\linewidth]{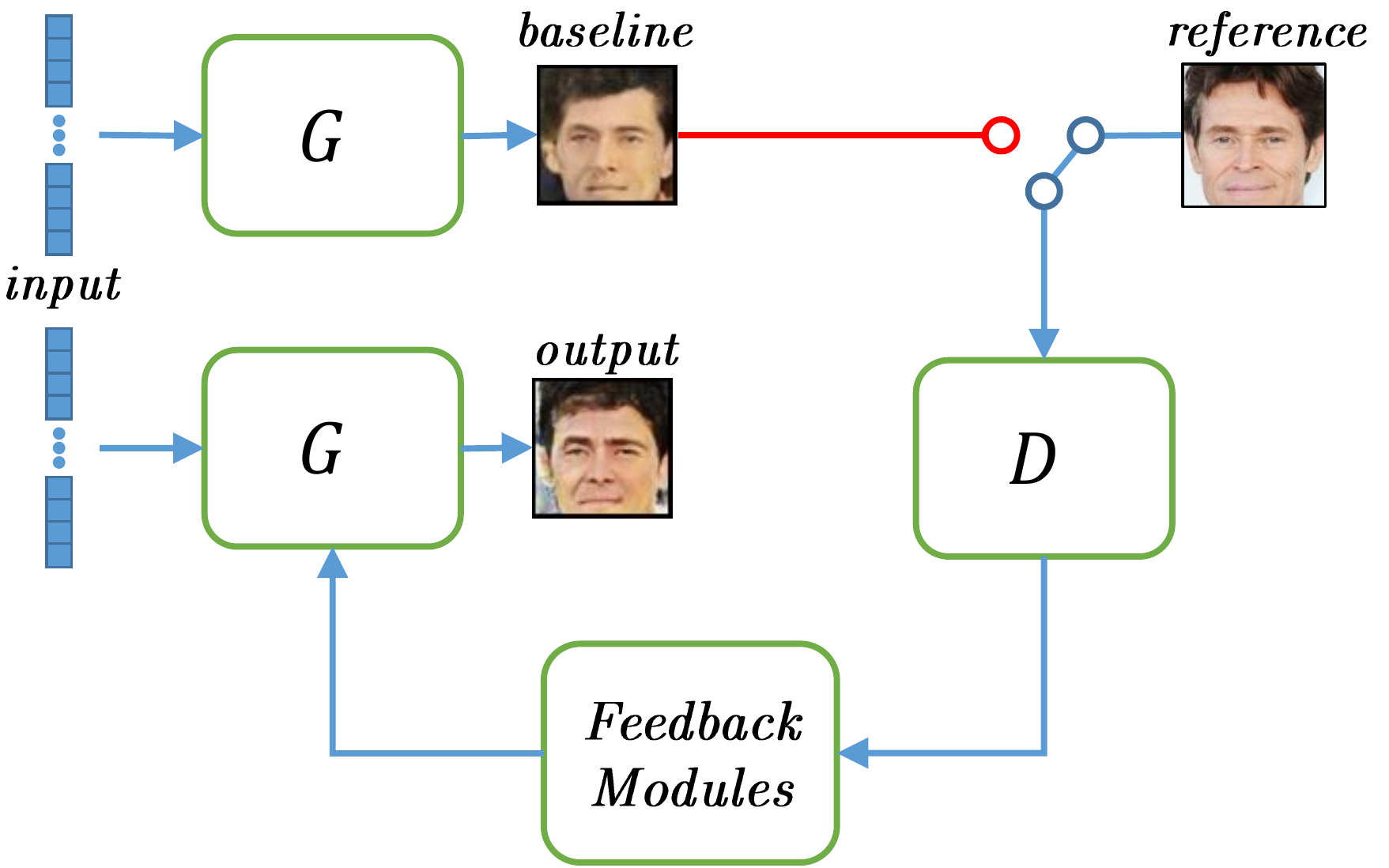}
  \caption{\textbf{Feedback switching pipeline:} instead of using AFL in its standard pipeline, we replace the input to the discriminator with a reference image, this allows us to control the similarity of the generated image to the reference image. }
  \label{fig:reference_pipe}
\end{figure}

Another experiment we performed was in order to further study the utility of the information encompassed in the discriminator.
To do that we investigated what happens when a feedback of an image is replaced by a feedback from another image, see Figure~\ref{fig:reference_pipe} for illustration.
This means that the feedback modules that correct the generation of image $A$ receive their input from a discriminator that is fed with another image $B$, namely:
\begin{equation}
    y_1 = G\left(x_A,\alpha\cdot\E(B)\right)
\end{equation}

Interestingly, this experiment's results show that with such a setup, the feedback modules ``correct'' the output image in the direction of the reference image.
Rather than correcting artifacts, the AFL modifies the content of the image to match the information it gets from the discriminator.
This shows that the features extracted by the discriminator are powerful. 
As can be seen in Figure~\ref{fig:referece_feedback}, when we set the hyper-parameter $\alpha$ to a higher value, we get an image with higher similarity to the reference image, and vice versa.
Furthermore, the faces suffer from fewer artifacts than those generated by the baseline.
Potentially, with such a framework we can control the randomly generated images of GANs.

Note that a similar experiment was presented previously with CIFAR-10 dataset, however since the dataset includes images of different classes of objects, modifying the content of the generated image from class A to be more similar to an object from class B, may end up with an unidentified object, which caused a drop in the IS score.

\input{fig_reference_fb.tex}

%% file: fig_FIDvsAlpha.tex
\begin{figure}
\centering
\setlength{\tabcolsep}{.0em}
\begin{tabular}{cccccc ccccc}
    $0.0$ & $0.1$ & $0.2$ & $0.3$ & $0.4$ & $0.5$ & $0.6$ & $0.7$ & $0.8$ & $0.9$ & $1.0$\\
    \includegraphics[width=.09\linewidth]{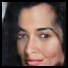}&
    \includegraphics[width=.09\linewidth]{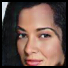}&
    \includegraphics[width=.09\linewidth]{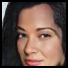}&
    \includegraphics[width=.09\linewidth]{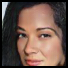}&
    \includegraphics[width=.09\linewidth]{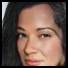}&
    \includegraphics[width=.09\linewidth]{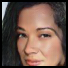}&
    \includegraphics[width=.09\linewidth]{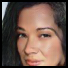}&
    \includegraphics[width=.09\linewidth]{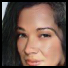}&
    \includegraphics[width=.09\linewidth]{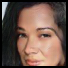}&
    \includegraphics[width=.09\linewidth]{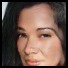}&
    \includegraphics[width=.09\linewidth]{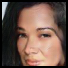}\\ 
\end{tabular}
\vspace{0.2cm}
\includegraphics[width=0.99\linewidth]{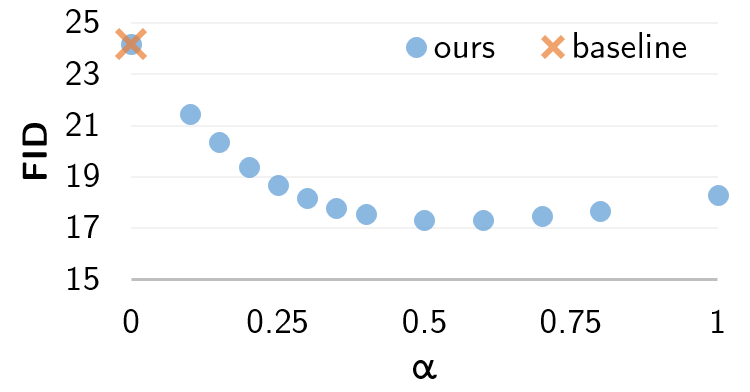}
\label{fig:alpha_images}
\vspace{-0.2cm}
\caption{\textbf{Quantitative evaluation on CelebA:} The graph shows the mean FID over 50K samples for multiple $\alpha$ value. It can be seen that AFL always improves the score. }
\label{fig:FIDvsAlpha}
\end{figure}

%% file: fig_disabling_modules.tex
\begin{figure}
\centering
\begin{tabular}{cccccc}
    \includegraphics[width=.13\linewidth]{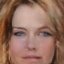}&
    \includegraphics[width=.13\linewidth]{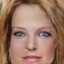} &
    \includegraphics[width=.13\linewidth]{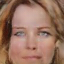} &
    \includegraphics[width=.13\linewidth]{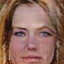} &
    \includegraphics[width=.13\linewidth]{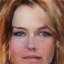} &
    \includegraphics[width=.13\linewidth]{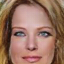}  \\
    
    \includegraphics[width=.13\linewidth]{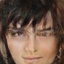}&
    \includegraphics[width=.13\linewidth]{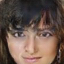} &
    \includegraphics[width=.13\linewidth]{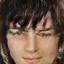} &
    \includegraphics[width=.13\linewidth]{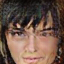} &
    \includegraphics[width=.13\linewidth]{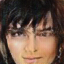} &
    \includegraphics[width=.13\linewidth]{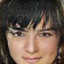}  \\
    
    \includegraphics[width=.13\linewidth]{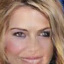}&
    \includegraphics[width=.13\linewidth]{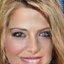} &
    \includegraphics[width=.13\linewidth]{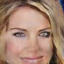} &
    \includegraphics[width=.13\linewidth]{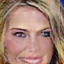} &
    \includegraphics[width=.13\linewidth]{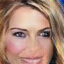} &
    \includegraphics[width=.13\linewidth]{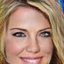}  \\
    
    \includegraphics[width=.13\linewidth]{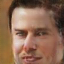}&
    \includegraphics[width=.13\linewidth]{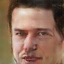} &
    \includegraphics[width=.13\linewidth]{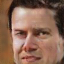} &
    \includegraphics[width=.13\linewidth]{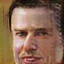} &
    \includegraphics[width=.13\linewidth]{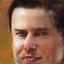} &
    \includegraphics[width=.13\linewidth]{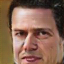} \\
    none & $1^{st}$ & $2^{nd}$ & $3^{rd}$  & $4^{th}$ & all
\end{tabular}
\caption{{\textbf{Analysis of module contribution:}} Face generation results while disabling all modules except the one specified. Shallow modules have the most impact on the result, while deep modules affect only colors and fine details.}
\label{fig:disabling_modules_exp}
\end{figure}

%% file: fig_reference_fb.tex
\begin{figure}
\centering
\setlength{\tabcolsep}{.1em}
\begin{tabular}{cc}
\textbf{generated samples for different $\mathbf{\alpha}$ values} & \textbf{reference}\\
\renewcommand{\arraystretch}{0.5}
\setlength{\tabcolsep}{.1em}
\begin{tabular}{cccccc}
    \includegraphics[width=.12\linewidth]{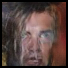}&
    \includegraphics[width=.12\linewidth]{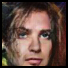}&
    \includegraphics[width=.12\linewidth]{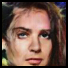}&
    \includegraphics[width=.12\linewidth]{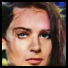}&
    \includegraphics[width=.12\linewidth]{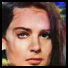}&
    \includegraphics[width=.12\linewidth]{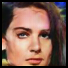}\\
    
    \includegraphics[width=.12\linewidth]{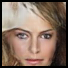}&
    \includegraphics[width=.12\linewidth]{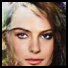}&
    \includegraphics[width=.12\linewidth]{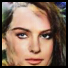}&
    \includegraphics[width=.12\linewidth]{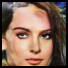}&
    \includegraphics[width=.12\linewidth]{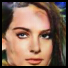}&
    \includegraphics[width=.12\linewidth]{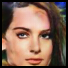}\\
    
    \includegraphics[width=.12\linewidth]{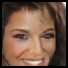}&
    \includegraphics[width=.12\linewidth]{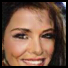}&
    \includegraphics[width=.12\linewidth]{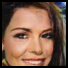}&
    \includegraphics[width=.12\linewidth]{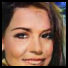}&
    \includegraphics[width=.12\linewidth]{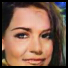}&
    \includegraphics[width=.12\linewidth]{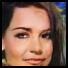}\\
    
    \includegraphics[width=.12\linewidth]{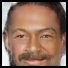}&
    \includegraphics[width=.12\linewidth]{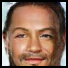}&
    \includegraphics[width=.12\linewidth]{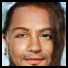}&
    \includegraphics[width=.12\linewidth]{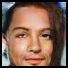}&
    \includegraphics[width=.12\linewidth]{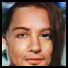}&
    \includegraphics[width=.12\linewidth]{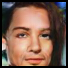}\\
    
    \includegraphics[width=.12\linewidth]{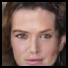}&
    \includegraphics[width=.12\linewidth]{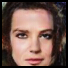}&
    \includegraphics[width=.12\linewidth]{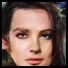}&
    \includegraphics[width=.12\linewidth]{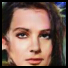}&
    \includegraphics[width=.12\linewidth]{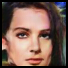}&
    \includegraphics[width=.12\linewidth]{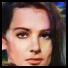}\\
    
    \includegraphics[width=.12\linewidth]{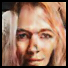}&
    \includegraphics[width=.12\linewidth]{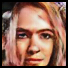}&
    \includegraphics[width=.12\linewidth]{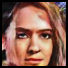}&
    \includegraphics[width=.12\linewidth]{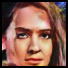}&
    \includegraphics[width=.12\linewidth]{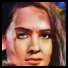}&
    \includegraphics[width=.12\linewidth]{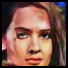}\\
    
    \includegraphics[width=.12\linewidth]{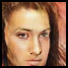}&
    \includegraphics[width=.12\linewidth]{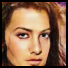}&
    \includegraphics[width=.12\linewidth]{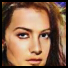}&
    \includegraphics[width=.12\linewidth]{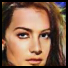}&
    \includegraphics[width=.12\linewidth]{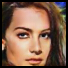}&
    \includegraphics[width=.12\linewidth]{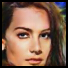}\\
\end{tabular}& 
\includegraphics[width=.2\linewidth]{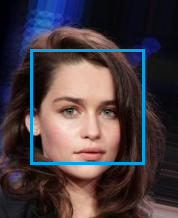}\\
\\
\setlength{\tabcolsep}{.1em}
\renewcommand{\arraystretch}{0.5}
\begin{tabular}{cccccc}
    \includegraphics[width=.12\linewidth]{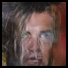}&
    \includegraphics[width=.12\linewidth]{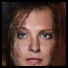}&
    \includegraphics[width=.12\linewidth]{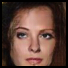}&
    \includegraphics[width=.12\linewidth]{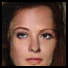}&
    \includegraphics[width=.12\linewidth]{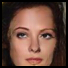}&
    \includegraphics[width=.12\linewidth]{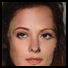}\\
    
    \includegraphics[width=.12\linewidth]{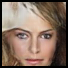}&
    \includegraphics[width=.12\linewidth]{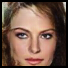}&
    \includegraphics[width=.12\linewidth]{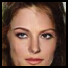}&
    \includegraphics[width=.12\linewidth]{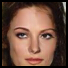}&
    \includegraphics[width=.12\linewidth]{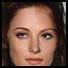}&
    \includegraphics[width=.12\linewidth]{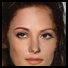}\\

    \includegraphics[width=.12\linewidth]{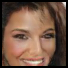}&
    \includegraphics[width=.12\linewidth]{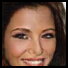}&
    \includegraphics[width=.12\linewidth]{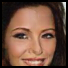}&
    \includegraphics[width=.12\linewidth]{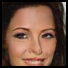}&
    \includegraphics[width=.12\linewidth]{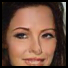}&
    \includegraphics[width=.12\linewidth]{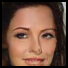}\\
    
    \includegraphics[width=.12\linewidth]{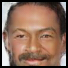}&
    \includegraphics[width=.12\linewidth]{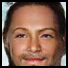}&
    \includegraphics[width=.12\linewidth]{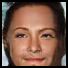}&
    \includegraphics[width=.12\linewidth]{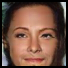}&
    \includegraphics[width=.12\linewidth]{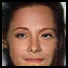}&
    \includegraphics[width=.12\linewidth]{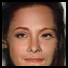}\\
    
    \includegraphics[width=.12\linewidth]{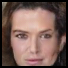}&
    \includegraphics[width=.12\linewidth]{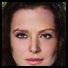}&
    \includegraphics[width=.12\linewidth]{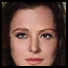}&
    \includegraphics[width=.12\linewidth]{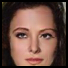}&
    \includegraphics[width=.12\linewidth]{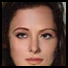}&
    \includegraphics[width=.12\linewidth]{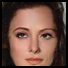}\\
    
    \includegraphics[width=.12\linewidth]{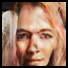}&
    \includegraphics[width=.12\linewidth]{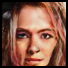}&
    \includegraphics[width=.12\linewidth]{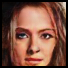}&
    \includegraphics[width=.12\linewidth]{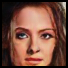}&
    \includegraphics[width=.12\linewidth]{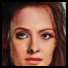}&
    \includegraphics[width=.12\linewidth]{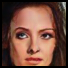}\\
    
    \includegraphics[width=.12\linewidth]{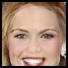}&
    \includegraphics[width=.12\linewidth]{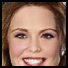}&
    \includegraphics[width=.12\linewidth]{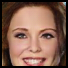}&
    \includegraphics[width=.12\linewidth]{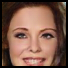}&
    \includegraphics[width=.12\linewidth]{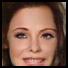}&
    \includegraphics[width=.12\linewidth]{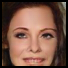}\\
    \\
    $0.0$\cite{radford2015unsupervised} & $0.1$ & $0.2$ & $0.3$ & $0.4$ & $0.5$ \\
\end{tabular}& 
\includegraphics[width=.2\linewidth]{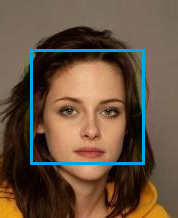}
\end{tabular}
\vspace{-0.2cm}
\caption{\textbf{Generation with reference:} Results of using the feedback-switching-pipeline of Figure~\ref{fig:reference_pipe}.
The feedback modules make the generated image similar to the reference one, and with fewer artifacts. This makes an interesting approach for controlled face generation. First column is DCGAN~\cite{radford2015unsupervised} baseline.}
\label{fig:referece_feedback}
\end{figure}

%% file: SR.tex
\input{fig_SR_examples.tex}

\subsection{Super Resolution: AFL in conditional GAN}
\label{sec:SR}

Our experiments so far showed the benefits of AFL in concert with a variety of GANs for unconditional image generation.
In all of these cases the objective is simple, consisting of only an adversarial loss, and the structure of generator and discriminator are almost symmetric.
This enabled straight-forward feedback loop from each discriminator layer to the equally-sized generator layer. 
Our next goal is to show the generality of AFL to more complex objectives and to non symmetric generator-discriminator structures. 
Specifically, we chose the problem of $\times4$ super-resolution, where we hope that AFL will correct artifacts and fine details produced by the baseline network, e.g., the blurriness and checker-board patterns shown in Figure~\ref{fig:SR_figure} (baseline). 



Our solution is based upon the top ranked method in PIRM super-resolution challenge~\cite{blau20182018}, named ESRGAN~\cite{wang2018esrgan}\footnote{Used the official implementation \href{https://github.com/xinntao/ESRGAN}{github/xinntao/ESRGAN}}, which builds upon SRGAN~\cite{ledig2017photo}, the first to use GANs for super-resolution. 
The key ideas behind ESRGAN are to use dense residual-in-residual as the main building block, optimize via relativistic GAN~\cite{jolicoeur2018relativistic}, abandon batch-normalization and use the features before activation in the perceptual loss~\cite{johnson2016perceptual}.
These design decisions make the architecture of the generator more complicated than the classic architectures used in our previous experiments. Moreover, it raises a new challenge since the architectures of the generator and the discriminator differ fundamentally. 

Our proposed AFL approach is generic enough to be easily adapted to such complex architectures. 
In this specific case we make the following adjustments:
we use 4 dual-input feedback modules, spread along the generator residual blocks (see supplementary), each with three convolutional layers. 
Since ESRGAN avoids batch-normalization in the generator, we also remove this from the feedback modules.
Since the layers of the generator and discriminator are not of the same dimension we upscale the discriminator features to match the size of the generator features.

\begin{table}[t]
    \centering
    \begin{tabu}{@{}ccccc@{}}
      \toprule
	\textbf{Method} & \phantom{ab} &  \textbf{PI~\cite{blau20182018}} & \phantom{ab} & \textbf{RMSE}\\
    \midrule
         ESRGAN && $2.433$ && $16.34$ \\
         ESRGAN+AFL (ours) && $\textbf{2.135}$ && $\textbf{14.72}$\\
    \bottomrule
    \end{tabu}
    \vspace{-0.1cm}
    \caption{\textbf{Super-resolution results:} The table reports the mean scores over PIRM test set~\cite{blau20182018}. RMSE measures similarity to the ground-truth, while PI~\cite{blau20182018} measures non-reference perceptual quality. It can be clearly seen that adding AFL leads to an improvement on both scores.}
    \label{tab:sr_table}
\end{table}

We follow the training process of the original ESRGAN~\cite{wang2018esrgan}.
The network and the feedback modules are trained with DIV2K dataset~\cite{timofte2017ntire}, Adam optimizer and with an objective combining three loss terms: relativistic GAN~\cite{jolicoeur2018relativistic}, perceptual loss~\cite{johnson2016perceptual} and L1. 
At test time we set $\alpha=1$.

We evaluate the contribution of AFL via two measures RMSE and the Perceptual Index [PI], as suggested in PIRM~\cite{blau20182018}.
The scores reported in Table~\ref{tab:sr_table} show that adding AFL improves both measures.
This supports the broad applicability of our proposed feedback approach. 

Qualitative assessment of the results is provided in Figure~\ref{fig:SR_figure}. It can be seen that AFL reduces artifacts, corrects textures, and improves the realistic appearance of the generated high-resolution image.

%% file: fig_SR_examples.tex
\begin{figure}
\centering
\setlength{\tabcolsep}{.0em}
\begin{tabular}{cc}
    \includegraphics[width=.5\linewidth]{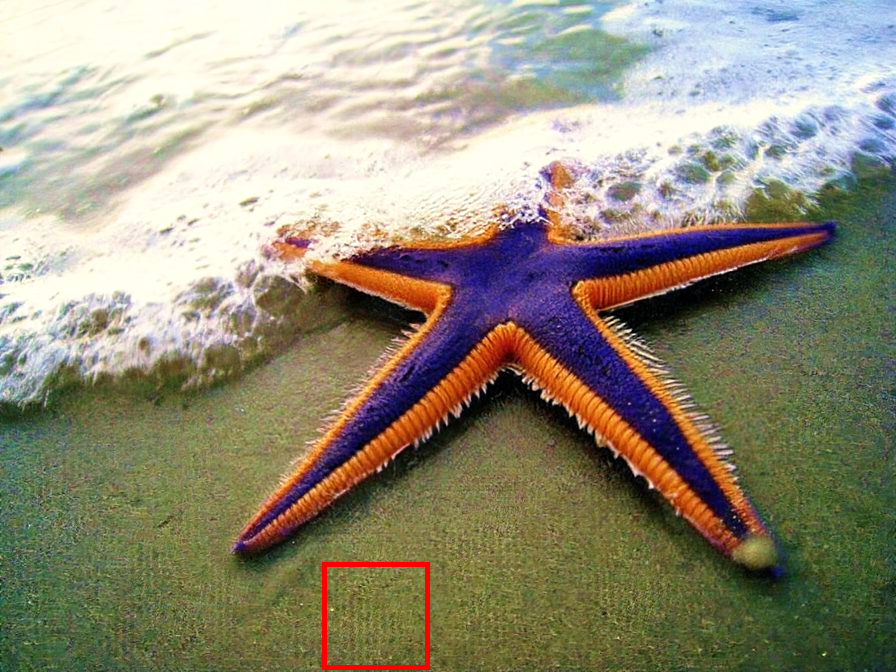}&
    \includegraphics[width=.5\linewidth]{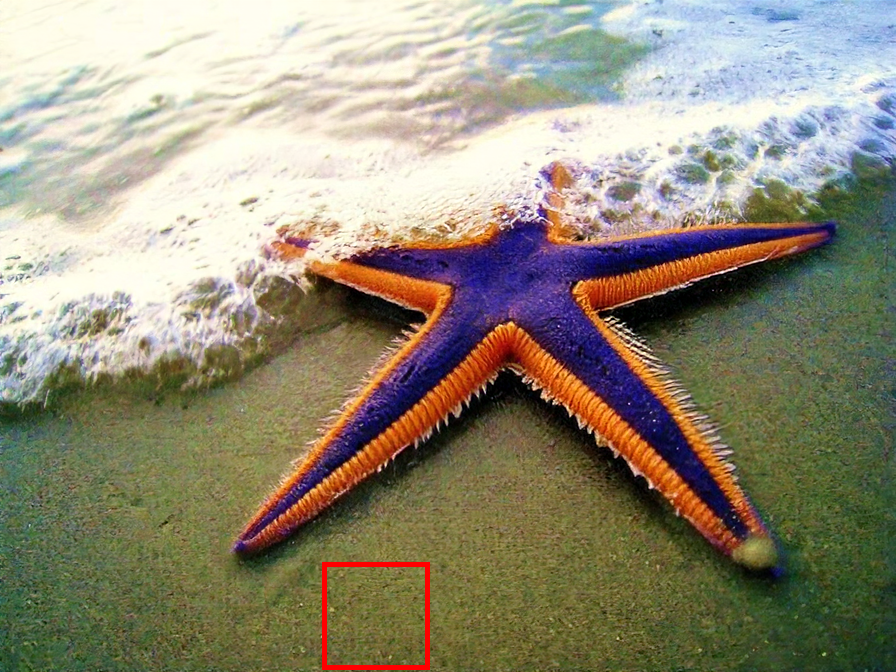}\\
    \includegraphics[width=.5\linewidth]{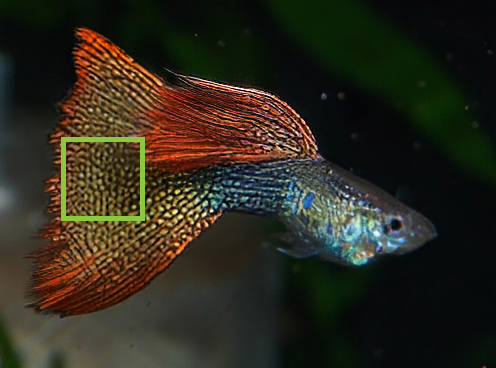}&
    \includegraphics[width=.5\linewidth]{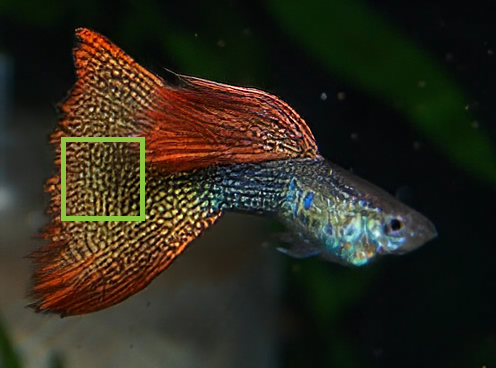}\\
    baseline & ours \\
    \end{tabular}
    
    \begin{tabular}{ccccc}
    \includegraphics[width=.20\linewidth]{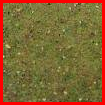}&
    \includegraphics[width=.20\linewidth]{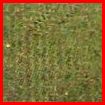}&
    \includegraphics[width=.20\linewidth]{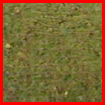}&
    \includegraphics[width=.20\linewidth]{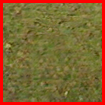}&
    \includegraphics[width=.20\linewidth]{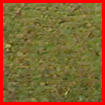}\\
    \includegraphics[width=.20\linewidth]{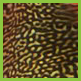}&
    \includegraphics[width=.20\linewidth]{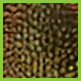}&
    \includegraphics[width=.20\linewidth]{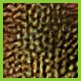}&
    \includegraphics[width=.20\linewidth]{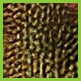}&
    \includegraphics[width=.20\linewidth]{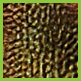}\\
    GT & baseline & $1^{st}$ iter  & $2^{nd}$ iter  & $3^{rd}$ iter  \\
    \end{tabular}
    \vspace{-0.2cm}
\caption{our feedback loop removes artifacts and sharpen the image }
\label{fig:SR_figure}
\end{figure}

%% file: conclusion.tex
\section{Conclusion}
In this work we proposed a novel idea of applying a feedback loop at test-time as part of the GAN framework, thus utilizing the discriminator not only for training. 
We showed via several applications, with various architectures and setups, that such an approach improves the quality of the generated images. 

Our approach has further potential to open-up new opportunities in image generation.
Specifically, controlled image generation using a reference image could have broad applications.
To leverage this we have built an interactive user interface, where a user can tune the impact of the feedback modules (by modifying $\alpha$). We are now exploring the benefits of this user-interface, but preliminary experiments suggest that in many cases user-specific and image-specific tuning could be a good option. 
